\documentclass[letterpaper, 10 pt, conference]{ieeeconf}  

\IEEEoverridecommandlockouts                              

\usepackage{graphicx}
\usepackage{amsmath,amssymb} 
\usepackage{color}
\usepackage{multirow}
\usepackage{flushend}

\usepackage{subcaption}
\usepackage{algorithm, algpseudocode}

\usepackage{array}
\usepackage{paralist}
\usepackage{pgfplotstable}
\usepackage{multicol}

\renewcommand{\baselinestretch}{0.95}

\DeclareRobustCommand{\iou}{IoU }
\DeclareRobustCommand{\iouns}{IoU}

\DeclareRobustCommand{\E}{\mathbb{E}}

\DeclareMathOperator*{\argmax}{\arg\!\max}

\newtheorem{theorem}{Theorem}

\newtheorem{colExpErrorGrowerBound}{Corollary }

\makeatletter
\def\bstctlcite{\@ifnextchar[{\@bstctlcite}{\@bstctl
cite[@auxout]}}
\def\bstctlcite[#1]#2{\@bsphack
\@for\@citeb:=#2\do{%
\edef\@citeb{\expandafter\@firstofone\@citeb}%
\if@filesw\immediate\write\csname #1\endcsname{\s
tring\citation{\@citeb}}\fi}%
\@esphack}
\makeatother

\title{\LARGE \bf
Visual Chunking: A List Prediction Framework for Region-based Object Detection} 
\author{Nicholas Rhinehart, Jiaji Zhou, Martial Hebert, and J. Andrew Bagnell
           \\The Robotics Institute, Carnegie Mellon University 
            \\ \{\texttt{nrhineha, jiajiz, hebert, dbagnell}\}\texttt{@cs.cmu.edu}
}

\begin{document}
\maketitle
\thispagestyle{empty}
\pagestyle{empty}

\begin{abstract} We consider detecting objects in an image by iteratively
selecting from a set of arbitrarily shaped candidate regions. Our generic approach, which we term
\textbf{visual chunking}, reasons about the locations of multiple object instances in an image
while expressively describing object boundaries. We design an
optimization criterion for measuring the performance of a list of such
detections as a natural extension to a common
per-instance metric. 
We present an efficient algorithm with provable performance for building a high-quality list of detections from any candidate set of region-based proposals. We also develop a simple class-specific algorithm to generate a candidate region instance in near-linear time in the number of low-level superpixels that outperforms other region generating methods. 
In order to make predictions on novel images at testing time without access to ground truth, we develop learning approaches to emulate these algorithms' behaviors.
We demonstrate that our new approach outperforms sophisticated baselines on benchmark datasets.

\tiny


\end{abstract}

\section{Introduction} \label{sec:introduction}

We consider the problem of object detection, where the goal is to
identify parts of an image corresponding to objects of a particular
semantic type, e.g. ``car''.
In recent years, machine learning-based approaches have become
de-rigueur for addressing this difficult problem; one classical approach
 is to transform the problem into one of binary
classification, either on bounding boxes \cite{felzenszwalb2010object,dalal2005histograms}, or regions.
Such approaches (see Section~\ref{sec:related} for a detailed
discussion) typically follow a two stage procedure: 
\begin{enumerate}
\item{generate independent proposals to provide coverage across object instances}
\item{improve precision and reduce redundancy by pruning out highly overlapping proposals}
\end{enumerate}

Intuitively, the first step returns a set of proposals with high recall and the
second step improves the precision. 
For the second step, traditional approaches rely on a
combination of thresholds and arbitration techniques like
Non-Max Suppression (NMS) to produce a final output. Such methods,
while remarkably effective at identifying sufficiently separated
objects, still have difficulty simultaneously detecting objects that
are close together or overlap while preventing multiple detections of the same object (see Fig.~\ref{fig:us_vs_them_all}). While we provide contributions to both stages, our focus is on formalizing and improving the second stage.


\newlength{\introtablewidth}
\setlength{\introtablewidth}{.15\textwidth}

\begin{figure}[!ht]
\centering
\includegraphics[width=.45\columnwidth]{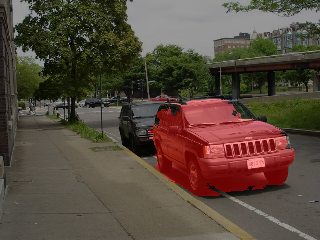}
\includegraphics[width=.45\columnwidth]{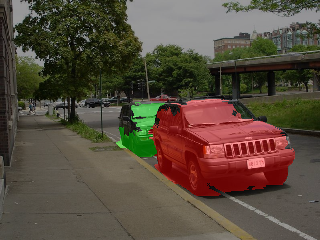}\\
\vspace{2pt}
\includegraphics[width=.24\columnwidth]{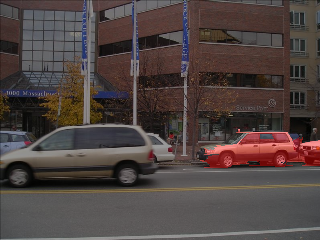}
\includegraphics[width=.24\columnwidth]{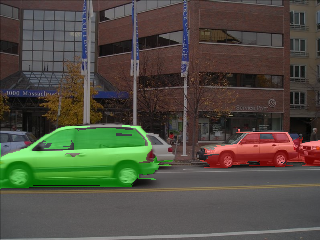}
\includegraphics[width=.24\columnwidth]{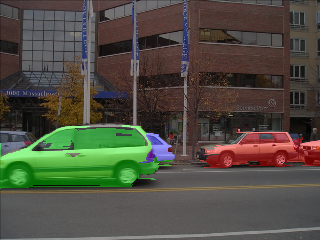}
\includegraphics[width=.24\columnwidth]{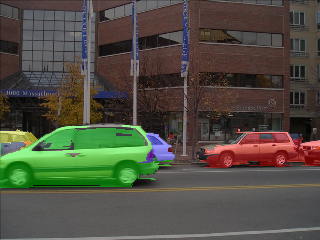}\\
\vspace{2pt}
\includegraphics[width=.32\columnwidth]{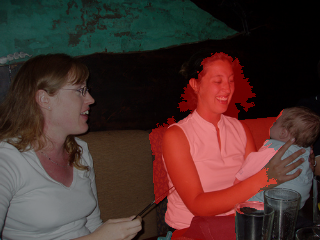}
\includegraphics[width=.32\columnwidth]{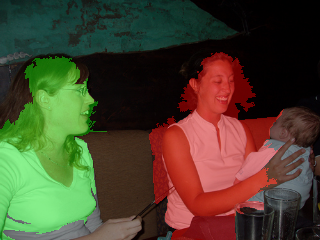}
\includegraphics[width=.32\columnwidth]{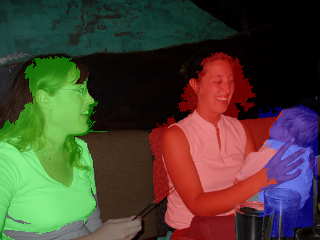}\\
\vspace{2pt} 
\includegraphics[width=.45\columnwidth]{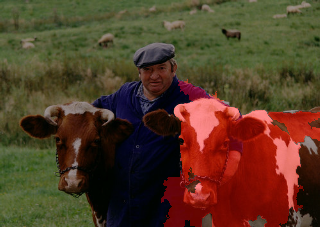}
\includegraphics[width=.45\columnwidth]{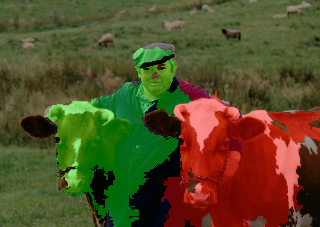}\\


\caption{Visual Chunking run on test data. The first prediction is shown in red, the second in green, the third in blue, and the fourth in yellow.} \label{fig:chunking_easy}
\hfill
\end{figure}

We formulate the objective of the second step as 
that of producing a diverse \emph{list} of detections in the
image. 
We propose an optimization criterion on this
list of detections as a natural extension of the intersection over union
metric (\iouns) (described in Section~\ref{sec:list_gt}), 
and develop an algorithm that targets this criterion. 
This approach uses recent work on building performance-bounded lists of
predictions \cite{dey2013contextual,ross2013learning}. 
Our algorithm shares information across all candidate detections to
build a list of detections, specifically exploiting this
information to perform well even when object instances are adjacent. Each
decision of appending to the list of detections is made with contextual
information from all previous detections. Importantly, our list prediction algorithm is agnostic to the source of candidate detections. This provides our approach with the ability to use \textit{any} candidate generating method as input for constructing a list.

Each candidate detection is treated as a union of superpixels with 
no adjacency constraints.  We call these unions ``chunks,'' inspired by a
well-known task in Natural Language Processing: ``chunking,'' which
involves grouping many words together into meaningful semantic instances / entities.
We use ``region'' to refer to a contiguous group of superpixels, and reserve
``chunk'' to refer to a group of superpixels corresponding to a single semantic instance.
The analogy is particularly apt when object instances 
are adjacent, as in Fig.~\ref{fig:chunking_easy}.

For the first step, we develop a class-specific supervised approach of region-based object
proposal by iteratively grouping superpixels produced by a low-level segmentation 
algorithm \cite{felzenszwalb2004efficient} to form chunks. This helps build a high-recall candidate set. 
This algorithm learns to ``grow'' by utilizing class-specific ground-truth labeling
by emulating an algorithm that optimizes a chunk's \iou score with an object, which we present in Algorithm~\ref{Grower}.
This strategy follows from imitation learning approaches \cite{daume2009search,ross2011reduction}.

Our technique for building the list of detections can be run for arbitrary list lengths, or budgets. This enables several use cases: building very short lists of highly confident object predictions (high precision), long lists of many candidate regions (high recall), and dynamic length lists tuned by some heuristic(s) (e.g., the highest predicted \iou score of the remaining candidates).

\section{Related Work} \label{sec:related}
Much work has been done in the combined areas of object detection and semantic
labeling. Object detection approaches often seek to place bounding
boxes around all instances of
objects \cite{felzenszwalb2010object,5459183}.
\cite{desai2011discriminative} casts the multi-class (and multi-instance) detection problem
as a structured prediction task instead of NMS as post
processing. However, the resulting detections are still bounding boxes.
   

Intermediate approaches deform the regions inside  the output of a
detector  to produce object segmentations
\cite{lempitsky2009image,monroy2012beyond,wang2001simplicity}, or,
conversely, adjust bounding boxes based on low-level features such as
boundaries, texture, and color
\cite{dai2012learning,mottaghi2012}. Again, these approaches refine
individual detections relying on the initial detector output. In
contrast, we attempt to find the best list of detections given a large
collection of candidate detections and regions. Closer to our
work, \cite{yang2012layered} proposes to use a deformable shape model
to represent object categories in order to extract region level object
detections. This approach reasons about occluders and overlapping
detection by using depth layering and is designed for one specific shape model for region-based representation, while our approach is agnostic to the source of region segments and detection boxes.

Direct region-based techniques, such as
\cite{leibe2008robust,gu2009recognition,uijlings2013selective}, use region-based
information to formulate detections, the produced detections are
bounding boxes, and detection performance is analyzed using individual
bounding box metrics. \cite{borenstein2002class} produces region-wise
segmentations, however they assume the existence of only one object in
each image. \cite{carreira2012object} produces multiple region-wise
segmentations, but contiguous and adjacent objects are not resolved,
and ignore inter-class context. Other region-based techniques are
segmentation algorithms that rely on combining low-level image features with class-specific models \cite{leibe2004combined,stella2002concurrent,levin2006learning,tu2005image}, control segmentation parameters from object detection \cite{kumar2005obj}, or use the box-level detections as features for segmentation \cite{gonfaus2010harmony,krahenbuhl2011efficient}. These approaches attempt to find regions that best agree with both the region segments and individual detections but do not explicitly deal with the problem of finding the most consistent {\em list} of detections as we do.

Semantic systems  such as \cite{heitz2008learning,gould2009decomposing,munoz-eccv-10,tighe2010superparsing} do produce region-level labels, which can be grouped into detections, however
\textit{there is no notion of separate detections}; connected
components of labeling are not grouped into their constituent object
instances. 
\cite{ion2011probabilistic} uses non-overlapping segmentation
proposals in its first stage, thus allowing, in principle, the handling of
multiple instances of the same class, without explicitly optimization for
multi-instance settings. 
Although the evaluation criteria in \cite{ion2011probabilistic}  focuses on per-class
overlap without accounting for multiple instances, the authors do note the possibility for multi-instance extension.
Combining semantic labeling with object detectors has been explored in
different ways. Several approaches were proposed to combine
pixel-level classification labels and box-level detections into a
single inference problem. For example,
\cite{ladicky2010,fidlerbottom,heitz2008cascaded,yao2012describing}
incorporate detections into a CRF model for semantic labeling. These
techniques attempt to generate a holistic representation of the scene
that combines objects and regions.  These approaches  rely on semantic
segmentation. Our approach, while incorporating semantic segmentation,
is agnostic to the input features, as well as to the source(s) from
which candidate detections are generated. 

Another group of approaches related to our work address the problem of generating proposals for regions or boxes that are likely to delineate objects, in a class-independent manner. The proposals can then be evaluated by a class-specific algorithm for object detection. They include, for example, generating regions by iterative superpixel grouping~\cite{levinshtein2010optimal,uijlings2013selective}, and ranking proposed regions~\cite{endres2010category} or boxes~\cite{alexe2010object,alexe2012measuring} based on a learned objectness score. In \cite{weiss2013scalpel}, the authors investigate an iterative, class-specific region generation procedure that incorporates class-specific models at different scales, and requires bounding boxes as input. Our generation method, in comparison, directly optimizes the instance-based \iou metric, and we provide worst-case and probabilistic performance bounds. All of these approaches are complementary to our work in that we can potentially use any of them as input to our candidate generation step, thus, we incorporate and compare to several of them in our experiments.

\section{Approach}\label{sec:approach}


Our task is to output a list of chunks, i.e., list of sets of superpixels as described in Section~\ref{sec:introduction}, with high intersection over union (\iouns) scores with each of the ground truth instances in the image. This metric is formalized in Section \ref{sec:list_gt}. We decompose the task into two parts:

\begin{compactitem}
\item{\textbf{Generation of a set of candidate chunks} containing some elements that cover individual object instances.}
\item{\textbf{Iterative construction of a list of chunks} by selecting
  from an arbitrarily generated set of candidate chunks so as to
  maximize a natural variant of intersection over union score for
  multiple object instances and multiple predictions.}
\end{compactitem}

In the second stage, the candidate chunks can be generated from any
algorithm, providing our method with the ability to augment our set of grown
candidates constructed by other means. We start by describing the
method by which we build lists of detections for the second stage, and first define a natural scoring function to evaluate any input
list of chunks given ground truth on the pixels corresponding to objects of interest in a scene. We provide an efficient greedy algorithm that is guaranteed to optimize this metric to within a constant factor given access to ground-truth and this arbitrary set of (potentially overlapping) candidate chunks. 

Our test-time approach, following recent work in structured
prediction \cite{ross2013learning,daume2009search}, is to learn to
emulate the sequential greedy strategy. The result is a
predictor that takes a candidate set of chunks and iteratively builds
a list of chunks that are likely to overlap well with separate objects in the scene.

We do not place assumptions on the given candidate set of
chunks: the list predictor is agnostic to the way the candidate set of
chunks is generated. 
Such a set can be heuristically generated in many ways, \emph{e.g.}, those created from the baseline approaches described in Section~\ref{sec:experiments}. In Section~\ref{sec:single_grower}, we provide an algorithm designed to generate a candidate based on a fixed superpixel-based segmentation, and in Section~\ref{sec:growing_multiple_instances} extend this algorithm to the case of growing multiple chunks per images.



\subsection{Objective function and greedy optimization} \label{sec:list_gt}
We establish an objective function to evaluate the quality of any
list, and devise a greedy algorithm to approximately maximize this
objective function given access to the ground-truth. This will lead to
the development of learning algorithm that produces a prediction procedure that operates on novel images.

Given an image with ground truth instance set $G = \{ g_1,...,g_m \}$
and candidate chunk set $C = \{c_1,...,c_n\}$, 
our goal is to sequentially build a list of chunks out of $C$ so as to maximize the
sum of \iouns's with respect to ground truth instances. 
Denoting $L = ( c_i, c_j, \dots, c_{k} )$ as a size-$k$ list of chunks,
we first establish correspondences between candidate chunks and ground
truth instances to enable pairwise \iou computation.
Note that each $c_i$ is associated with at most one ground truth instance $g_i$, and each $g_i$ is associated with at
most one $c_i$. For analytic convenience, we augment $G$ with $k - m$
dummy ground truth instances $\tilde{g}$ to deal with the case in which
the length of the list is larger than the number of ground truth
instances ($|L| > |G|$). Every chunk $c$ has zero intersection with each $\tilde{g}$. Each feasible assignment corresponds to a
permutation $\tilde{L} = (c_{p_1}, c_{p_2}, \dots, c_{p_k} ) $ of
$L$, and the sum of \iou
scores for this permutation can be written as the following: $h(\tilde{L};G) = \sum_{i=1}^{k}\frac{|c_{p_i}\cap{g_i}|}{|c_{p_i}\cup{g_i}|}$.
It is natural to define the quality metric $f(L;G)$ of a list $L$ to be the sum of
\iou scores under the optimal assignment, i.e.,
$f(L;G) = \max_{\tilde{L}\in P(L)} h(\tilde{L}, G)$, where $P(L)$
denotes all permutations of $L$. With an abuse of notation, $L \subseteq C$ indicates all elements in $L$
       belong to $C$. Our goal during training is to
find list $L$ to maximize $f$:
\begin{equation}
        \arg\max_{L\subseteq C} f(L;G) =\arg\max_{L\subseteq
       C}\{\max_{\tilde{L}\in P(L)} h(\tilde{L}; G) \ \}. \label{list_obj_func} 
\end{equation}
This scoring metric, which is a natural generalization of the \iou metric common in segmentation and single instance detection \cite{Everingham10,arbelaez2012semantic}, encourages 
lists of a fixed length that contain chunks that are relevant and diverse in covering multiple ground truth instances.
Unfortunately, the metric as written down does not possess a clear combinatorial structure like modularity or submodularity 
that would beget easy optimizability.

Interestingly, however, Problem~(\ref{list_obj_func}) can be cast as
an equivalent maximum weighted bi-partite graph matching problem. 
This problem can be shown to be a submodular maximization
problem under matroid partition constraints, and a greedy
algorithm as shown in Algorithm~\ref{alg:ListGreedy} has
multiplicative performance guarantees ~\cite{fisher1978analysis}. 
In addition to these guarantees, such a
greedy algorithm is desirable as it is easily imitable at test time, 
and has a recursive solution: the $k + 1$ length list is exactly
the $k$ length list with the next greedily chosen item appended.
The greedy algorithm behaves as follows: at each iteration, it chooses the
chunk with the highest \iou with one of the remaining ground truth instances.
More precisely, a chunk's best overlap with each remaining ground
truth is defined as $y(c;G_{re}) = \max_{g \in G_{re}}\frac{|c \cap
  g|}{|c \cup g|}$ (the ``greedy marginal''), where $G_{re}$ is the set of remaining unpaired ground truth instances. At each step, the algorithm chooses the chunk with the highest $y(c;G_{re})$ value, appends it to the list ($L^{gr}$), and removes its associated ground truth from the set of remaining ground truth. This associated ground truth element is given by
$\pi_{gr}(c;G_{re}) = \arg\max_{g \in G_{re}} \frac{|c \cap g|}{|c \cup
  g|}$.



\begin{algorithm}
\caption{Greedy List Generation with Ground-Truth Access \label{alg:ListGreedy}}
\begin{algorithmic}
\State \textbf{Input:} Set of candidate chunks $C$, set of ground
truth instances $G$, size of predicted list $k$
\State \textbf{Output:} A near-optimal list $L^{gr}$ of chunks
\State $L^{gr} =  \emptyset$, $G_{re} = G$ \
\For{$i=1$ \textbf{to} $k$}
\State $c_i^{gr} = \arg\max_{c \in C} y(c;G_{re})$, $g_i^{gr} = \pi_{gr}(c;G_{re})$. \Comment{\small \textit{choose the highest scoring (chunk, GT) pair}}
\State $L^{gr} = L^{gr} \oplus c_i^{gr}$. \Comment{\small \textit{append the chunk to the list}}
\State $G_{re} = G_{re}\setminus g_i^{gr}$. \Comment{\small \textit{remove the associated GT}}
\EndFor
\State Return $L^{gr}$
\end{algorithmic}
\end{algorithm}

Critically, the greedy algorithm is recursive, meaning longer lists of predictions always include shorter lists, and is within a constant factor of optimal\footnote{Although Problem~(\ref{list_obj_func}) can be solved exactly, it requires knowledge
of the instances to be matched and does not possess a recursive structure that enables
simple creation of longer lists of prediction.}:
\begin{theorem}\label{thm:opt_grower_bound}
Let $L_{i}^{gr}$ be the list of the first $i$ elements in $L^{gr}$ and $L_{i}^*$ be the optimal solution of Problem (\ref{list_obj_func}) among size-$i$ lists
\begin{align}
f(L_{i}^{gr}; G) \geq \frac{1}{2} f(L_{i}^*), \forall i = 1,\cdots,k.
\end{align}
\end{theorem}
See the appendix for proof of Theorem~\ref{thm:opt_grower_bound}, which invokes results from \cite{fisher1978analysis}.
Theorem~\ref{thm:opt_grower_bound} implies that if we are given a
budget $|L|$ to build the list, then each $L_{i}^{gr}$ scores within a constant factor of the
optimal list among all lists of budget $i$, for $i=1, \dots,
|L|$. This is an important property for producing good predictions
earlier in the list and for producing the list of chunks rapidly. The empirical performance is usually much better than this bound suggests. 

\subsection{List prediction learning }  \label{sec:list_learning}
In essence, the greedy strategy is a sequential list prediction
process where at each round it maximizes the marginal benefit given
the previous list of predictions and ground truth
association. Maximization of the marginal benefit at each
position of the list yields chunks that have high \iou
with ground truth instances and minimal overlap with each
other. At test time, however, there is no access to the ground-truth. Therefore,
we take a learning approach to emulate the greedy algorithm. 
       We train a predictor to imitate Algorithm~\ref{alg:ListGreedy}, 
with the goal of preserving the ranking of all candidate chunks based 
on the greedy increments $y(c;G_{re})$. This predictor uses both information
about the current chunk and information about the currently built list
to inform its predictions. In our experiments, we train
random forests as our regressor with features as $\Phi(c,L)$ (each chunk's feature
is a function of itself and the currently built list, as described in Section~\ref{sec:features}), and regression targets
$y(c;G_{re})$ (the score for a chunk at each iteration is the greedy marginal, or
how much a chunk candidate covers a new object instance). This regression of ``region \iou'' is similar to that
explored in \cite{carreira2012cpmc}, except it is explicitly reasoning about multiple objects, as well as the current contents of the predicted list.
The prediction procedure is similar to the greedy list
generation as in Algorithm~\ref{alg:ListGreedy}, with the difference that
there is no access to the ground truth.

\subsection{Growing for a single instance}\label{sec:single_grower}
To generate a set of diverse chunks (output of stage 1 in the
detection process), we develop a class-specific algorithm that
``grows'' chunks via iterative addition of superpixels, with the
goal of producing diverse candidate detections that cover each
ground truth object instance. We first analyze the case where there is only a single object of interest $g$ in the image. We consider a chunk $c$ to be a union of superpixels $s$, i.e., $c = \cup_{i=1}^n \{s_i\}$. Let $R(c)$ denote the \iou score between $c$ and $g$.

To grow a chunk, Algorithm~\ref{Grower} starts with an empty chunk (no
superpixels), and adds single superpixels to the current chunk
sequentially. After each addition, the resulting chunk is copied and
added to the set of candidate chunks.
Let $\alpha_i =  \frac{|s_i \cap g|}{|s_i|}$ be the ratio of intersection area with ground truth to the size of a superpixel $s_i$. The set of chunks generated by the greedy algorithm described in Algorithm \ref{Grower} is guaranteed to contain the optimal chunk if the input predictor $\mathcal{G}$ returned the exact value of $\alpha_i$, i.e., $\hat{\alpha_i} = \alpha_i$. 

\begin{algorithm}
\caption{Single Instance Chunk Growing Algorithm }\label{Grower}
\begin{algorithmic}
\State \textbf{Input:} Set of superpixels $S$, grower predictor $\mathcal{G}$.
\State \textbf{Output:} A set of chunks, $C_{\mathcal{G}}$.
\State $c =  \emptyset$, $C_{\mathcal{G}} = \emptyset$
\State Sort elements in $S$ by decreasing order of $\hat{\alpha_i} = \mathcal{G}(s_i)$
\For{$i=1$ \textbf{to} $|S|$}
\State $c = c \cup \{s_i\}$, $C_{\mathcal{G}} = C_{\mathcal{G}} \cup \{c\}$
\EndFor
\State Return $C_{\mathcal{G}}$
\end{algorithmic}
\end{algorithm}

\begin{theorem}
Let $\mathcal{G}^*$ be an oracle growing predictor, i.e., $
\mathcal{G}^*(s_i) = \alpha_i =  \frac{ |s_i \cap g|}{|s_i|} $. The
output set of $C_{\mathcal{G}}$ from Algorithm~\ref{Grower} by setting
$\mathcal{G} = \mathcal{G}^*$ contains the best chunk given the set of superpixels $S$. 
\end{theorem}

See appendix for proof.
At testing time, we must give an estimate of $\alpha_i$, i.e.,
$\hat{\alpha_i}$. We train a random forest
regressor as our predictor $\mathcal{G}$ with features $\theta$ for estimation.
We analyze the performance of Algorithm~\ref{Grower} under approximation by relating the squared regression error of
$\mathcal{G}$ to the \iou score of the grown best chunk in
the chain. We note that the test-time performance depends on both the
size of the squared error and the number of predictions made. Notably,
the error bound has no explicit dependence on the area sizes of ground
truth object instances and images. See appendix for proof.

\begin{theorem}
Given a regressor $\mathcal{G}$ that achieves no worse than absolute error $\epsilon$ uniformly across all superpixels, let $c_{\mathcal{G}}^*$  be the best chunk in the predicted set $C_{\mathcal{G}}$. The \iou score of $c_{\mathcal{G}}^*$ is no worse than $2\epsilon$ of the \iou score of the optimal chunk $c^*$: $R(c_{\mathcal{G}}^*) > R(c^*) - 2\epsilon$.
\end{theorem}

\begin{colExpErrorGrowerBound}
Suppose regressor $\mathcal{G}$ has expected sq. error
$\delta$ over the distribution of superpixels, let $n$ be the number
of superpixels in the image, then we have for any $\eta \in (0,1)$, with probability $1 - \eta$:
$R(c_{\mathcal{G}}^*) > R(c^*) - 2\eta^{-1}\sqrt{n\delta}$.
\end{colExpErrorGrowerBound}

\subsection{Growing for multiple instances}\label{sec:growing_multiple_instances}
We run the growing algorithm more than once to cover multiple objects. Instead of making predictions
based solely on features of each individual superpixel, we augment the
information available to the predictor by including a feature of the
current grown chunk, $\theta(s_i, c)$ (see Section~\ref{sec:features} for more information about grower features). This yields predictors that prefer choosing superpixels in close proximity
to the currently growing chunk, and allows us not to explicitly encode
contiguity requirements, as objects may be partially occluded in a way
that renders them discontiguous. We also modify Algorithm \ref{Grower}
by ``seeding'' the chunks at a set of superpixel locations, $L$ (the
initialization step, $c = \emptyset$, becomes $c = \{s_i\}$ $\forall
s_i \in L$), and running the growing procedure on each of these seeds
separately. See appendix for the pseudo-code description modified from Algorithm~\ref{Grower}. 
In practice, we choose a seeding grid interval and a maximum chunk size cutoff, yielding $|C_{\mathcal{G}}| \sim 700$. In Figure \ref{fig:growing_figure_true}, we visualize the sequential growth of the best chunks for each object instance.

\newlength{\imagelength}
\setlength{\imagelength}{.10\textwidth}
\newlength{\multilength}
\setlength{\multilength}{.10\textwidth}


\begin{table}
\centering
\vspace{2pt}
\tabcolsep=0.02cm
 \begin{tabular}{m{\imagelength}m{\imagelength}m{\imagelength}m{\imagelength}}
 \includegraphics[width=\imagelength]{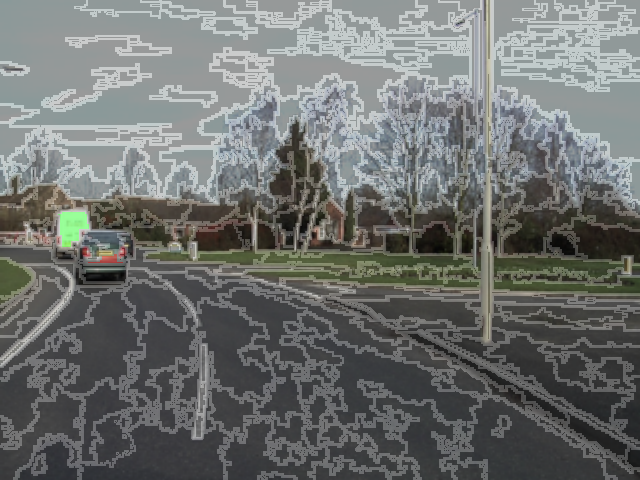}
 & \includegraphics[width=\imagelength]{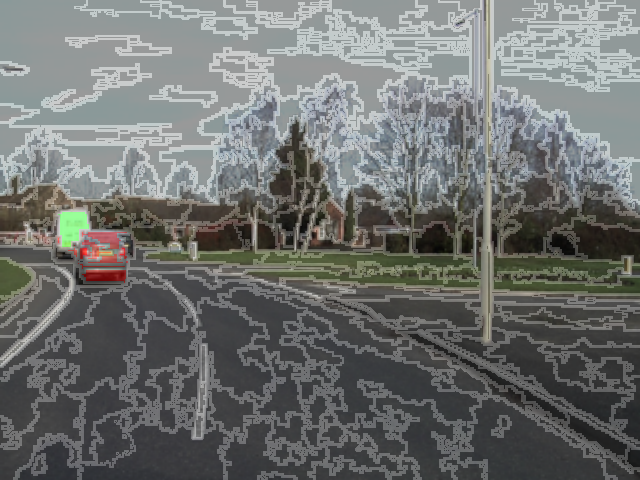}
 & \includegraphics[width=\imagelength]{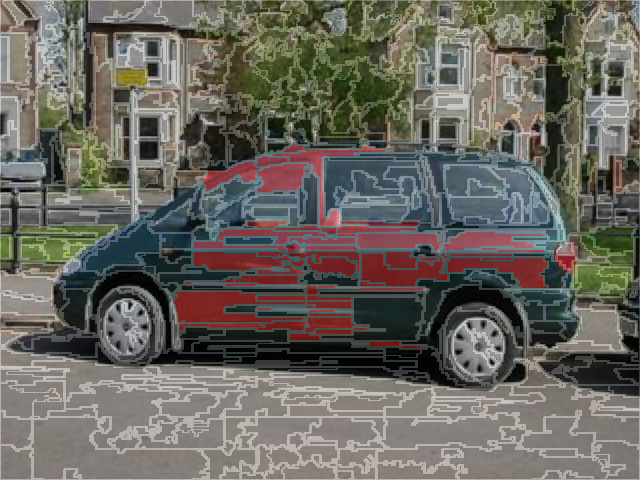}
 & \includegraphics[width=\imagelength]{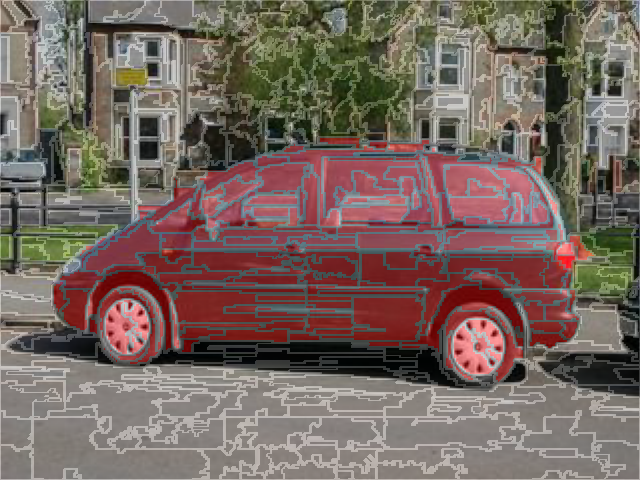}\\
\includegraphics[width=\multilength]{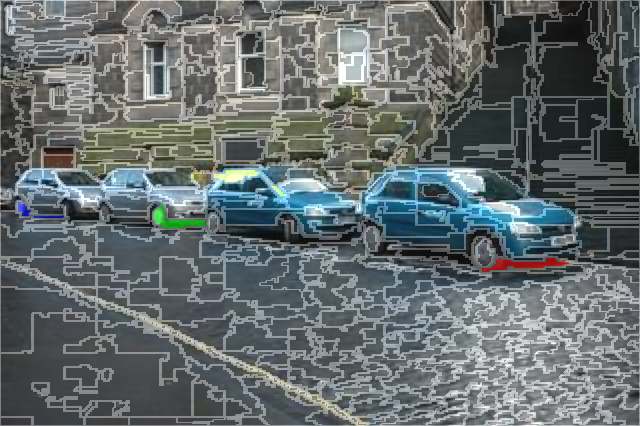} 
&\includegraphics[width=\multilength]{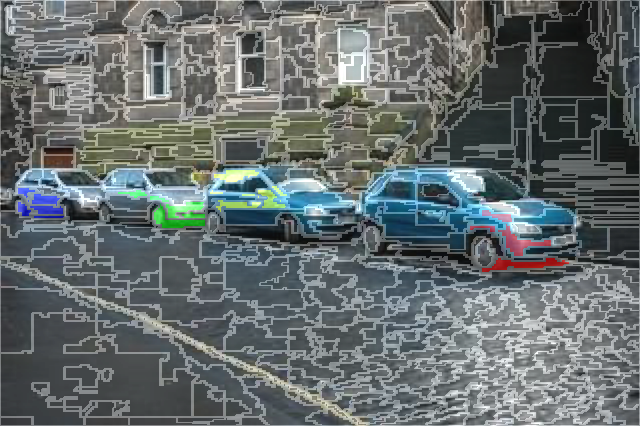} 
&\includegraphics[width=\multilength]{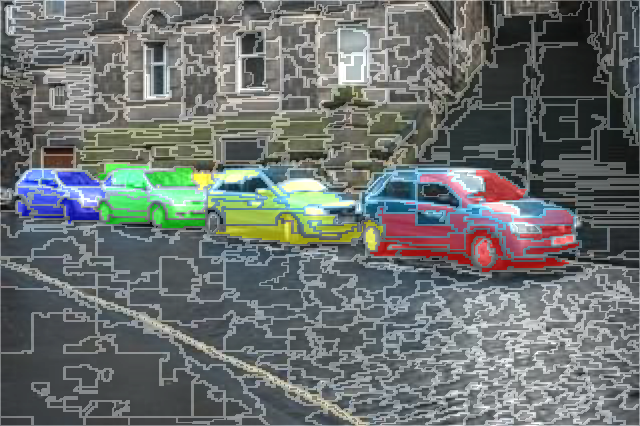} 
&\includegraphics[width=\multilength]{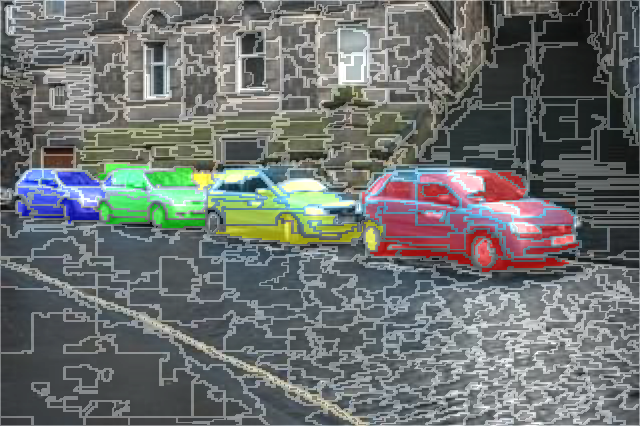} 
\end{tabular}
\small
\captionof{figure}{Selected images of the \emph{best} grown chunks for images with single and multiple objects. Each chunk grows independently of the others. Given the initial seed, superpixels are iteratively added to the growing chunk. The predictor greedily adds superpixels that it believes make the highest contribution to the overall class-specific \iou score of the currently growing chunk.}
\normalsize
\label{fig:growing_figure_true}
\end{table}

\section{Experiments} \label{sec:experiments}
We describe our experiments and features in the next two sections, and discuss the results of each experiment in their respective captions.

\newlength{\vocsinglewidth}
\setlength{\vocsinglewidth}{.10\textwidth}
\newlength{\doubleoffset}
\setlength{\doubleoffset}{2\normalbaselineskip}

\newlength{\lmsunwidth}
\setlength{\lmsunwidth}{.48\textwidth}

\begin{figure*}[t!]
  \centering
  \begin{minipage}[t]{\lmsunwidth}
    \includegraphics[width=\columnwidth]{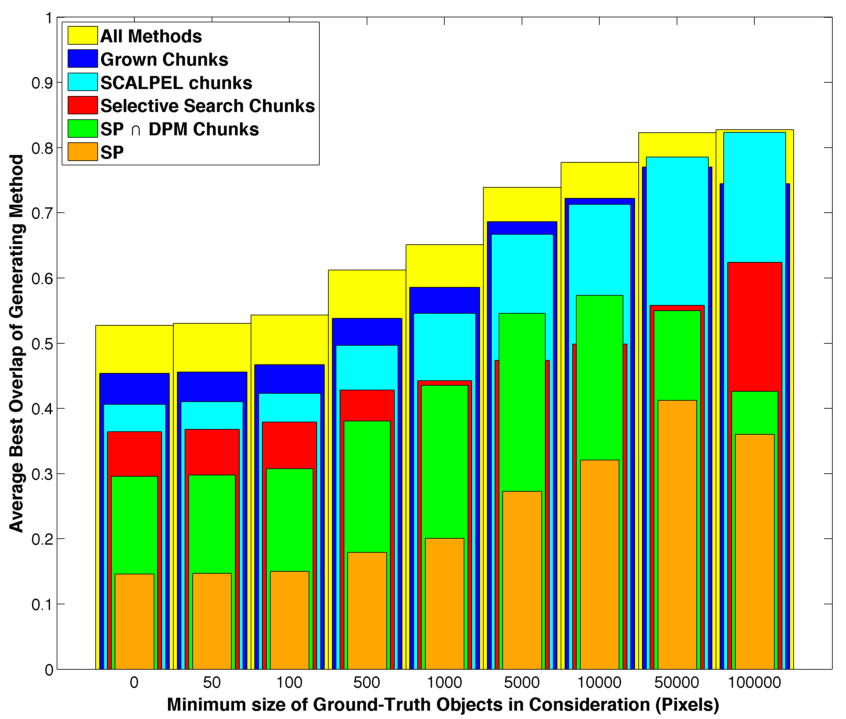}
    \caption{LM+Sun Adjacent Cars candidate quality, measured by Average Best Overlap (ABO). We find that our grown chunks generally outperform the Selective Search and SCALPEL methods. On average, SCALPEL generated $893$ regions per image, Selective Search generated $552$ regions per image, SP~$\cap$~DPM generated 8 chunks per image, and our grower generated $705$ chunks per image.} \label{fig:lmsun_abo_and_lp}
\end{minipage} \begin{minipage}[t]{\lmsunwidth}
            \includegraphics[width=\columnwidth]{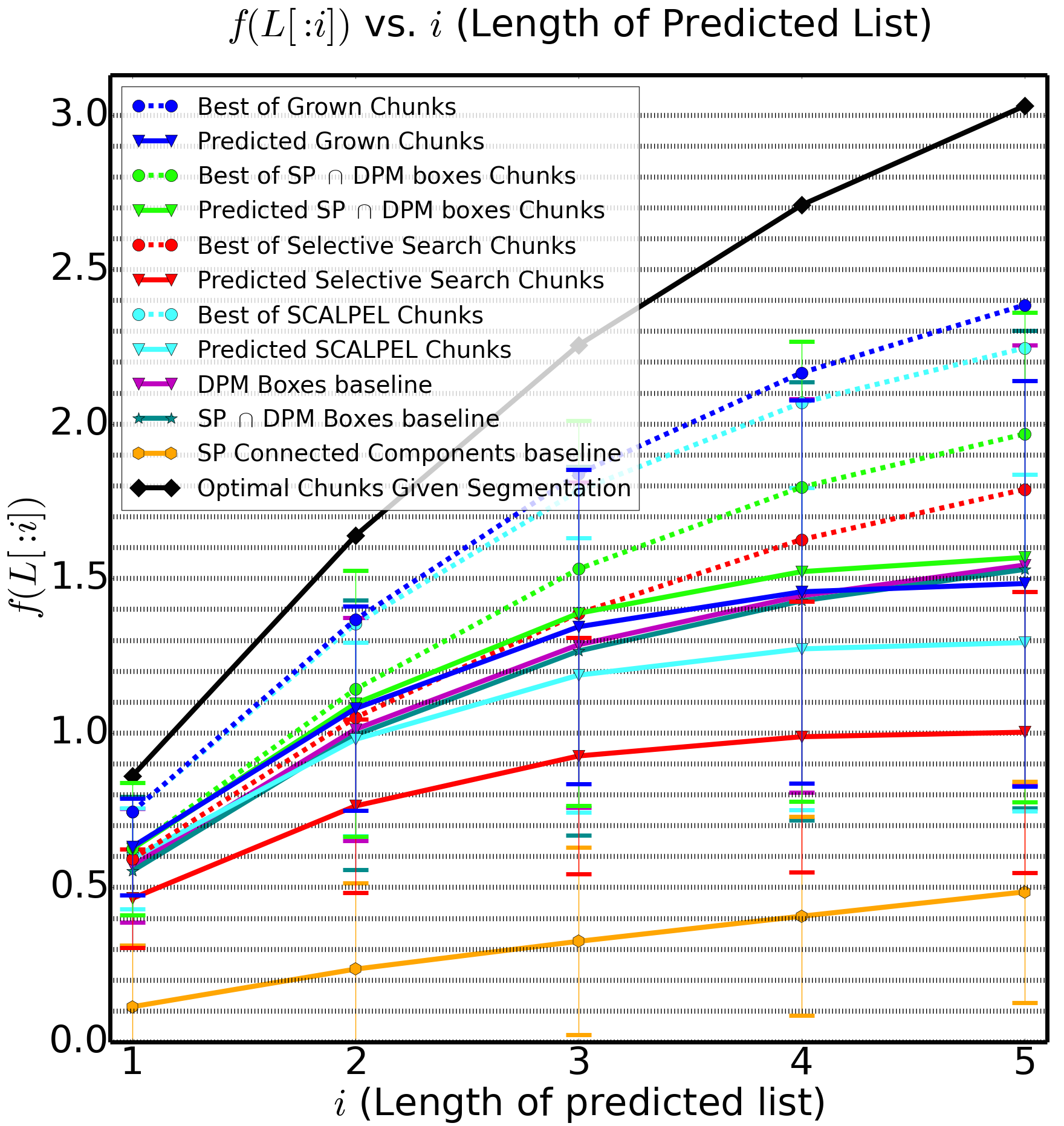}
    \caption{\footnotesize $f(L; G)$ performance of lists constructed from best candidates from each prediction pool (dashed lines) and predicted candidates (solid lines) on a 50/50 split of LM+Sun Adjacent Cars dataset. Our chunk generating method (dark blue) generates candidates of similar quality to that of SCALPEL (light blue). On this dataset, our DPM-based baselines (magenta and dark cyan) perform quite well, but the best performing list prediction method (green line) is our list predictor that uses the SP $\cap$ DPM chunks as the candidate pool, and essentially has learned how to reorder them. This demonstrates how our approach can utilize and improve different candidate sources.} \label{fig:sum_iu_lmsun}
\end{minipage}
\end{figure*}

\subsection{Datasets and baseline algorithms}
\begin{figure*}[t]
\tabcolsep=0.07cm
\centering
\begin{tabular}{cccccccc}
\multirow{2}{*}[\doubleoffset]{\includegraphics[width=\vocsinglewidth]{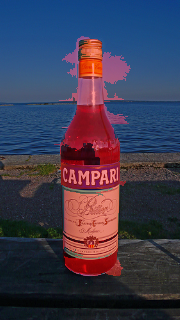}}  & \includegraphics[width=\vocsinglewidth]{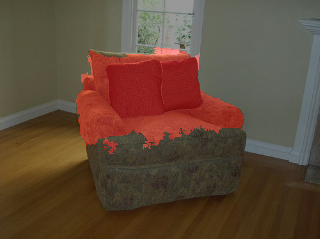}  &   \includegraphics[width=\vocsinglewidth]{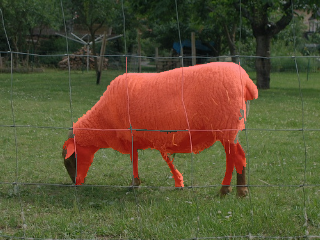} &   \includegraphics[width=\vocsinglewidth]{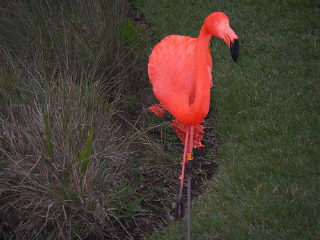} & \includegraphics[width=\vocsinglewidth]{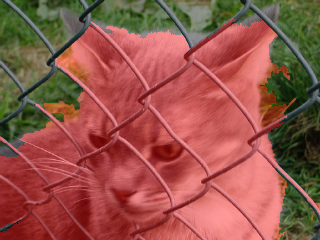} &   \includegraphics[width=\vocsinglewidth]{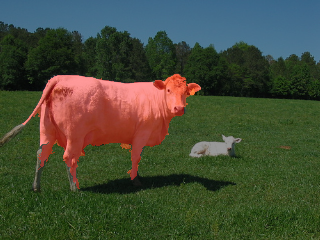} &   \includegraphics[width=\vocsinglewidth]{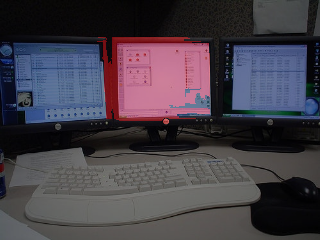} & \multirow{2}{*}[\doubleoffset]{\includegraphics[width=\vocsinglewidth]{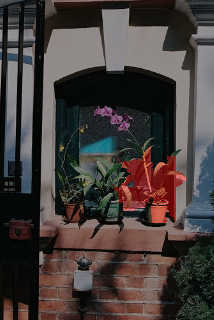}}\\   
 & \includegraphics[width=\vocsinglewidth]{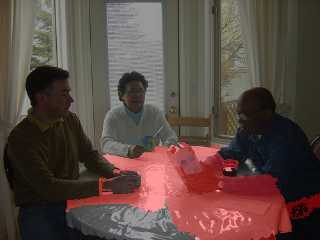} & \includegraphics[width=\vocsinglewidth]{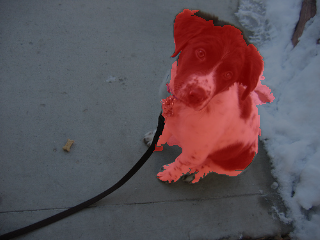} & \includegraphics[width=\vocsinglewidth]{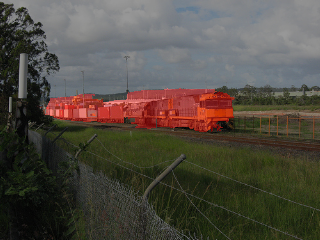} &   \includegraphics[width=\vocsinglewidth]{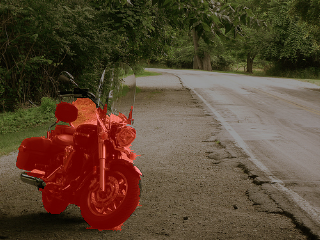} & \includegraphics[width=\vocsinglewidth]{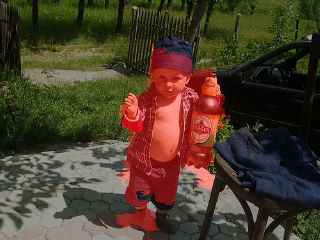}   &  \includegraphics[width=\vocsinglewidth]{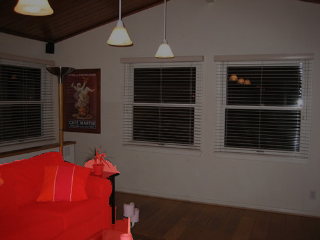} \\
 \includegraphics[width=\vocsinglewidth]{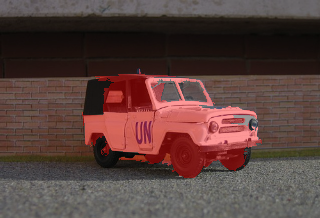} &    \includegraphics[width=\vocsinglewidth]{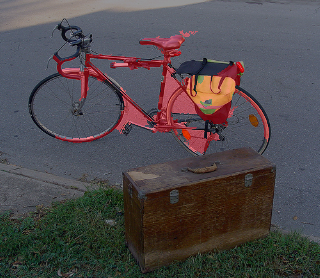}&   \includegraphics[width=\vocsinglewidth]{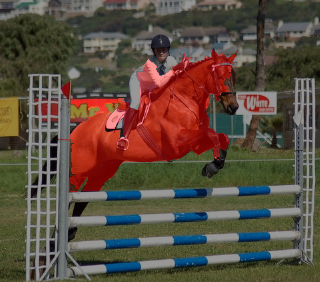} & \multirow{1}{*}[1.23\doubleoffset]{\includegraphics[width=\vocsinglewidth]{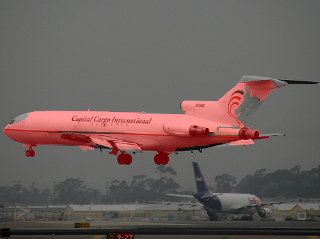}} &  \multirow{1}{*}[1.23\doubleoffset]{\includegraphics[width=\vocsinglewidth]{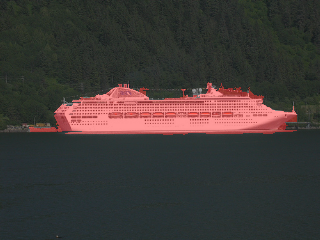}}   &   \multicolumn{2}{p{1em}}{\multirow{1}{*}[1.23\doubleoffset]{\includegraphics[width=1.5\vocsinglewidth]{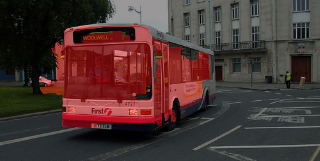}}} \\%
\end{tabular}\captionof{figure}{Example single predictions on PASCAL VOC2012}\label{fig:voc2012}

\tabcolsep=0.09cm
\captionof{table}{Average instance-based accuracy (a metric proposed by \cite{yang2012layered}), and first slot scores (corresponding to the average overlap of the first prediction in each image) for systems trained and tested on the standard PASCAL 2012 \texttt{train} and \texttt{val} sets. We find that the very small amount of co-occurring instance training data was not sufficient to enable our system to perform as well as it did in our other experiments on images with co-occurring instances. While \cite{yang2012layered} provides experimental results of average instance-based accuracy on PASCAL 2010, their results are confined to verified correct DPM detections, rendering a fair comparison difficult.} \label{tab:voc2012}
\begin{tabular}{|c|c|c|c|c|c|c|c|c|c|c|}
\hline
 & aeroplane & bicycle & bird & boat & bottle & bus  & car & cat & chair & cow \\
\hline
$\text{\text{inst}}_{\text{acc}}$ &.157 &.066    &.105 &.132 &   .079 &.228 &.097 & .155 & .071 & .211 \\
\hline
$f(L[0])$ &.521 &.148   &.375 &.335 & .186  &.439 &.190 &.445 &.141 & .494 \\
\hline
$f(L)$&  .530    &   .158  & .394 & .347 &.202&.509 & .195 & .461 & .171 & .581 \\
\hline
\end{tabular}
\vspace*{-5pt}
\begin{tabular}{|c|c|c|c|c|c|c|c|c|c|c|}
\multicolumn{11}{l}{}\\[-5pt]
\hline
 & diningtable & dog  & horse & motorbike & person & pottedplant & sheep & sofa & train & tvmonitor \\
\hline
$\text{\text{inst}}_{\text{acc}}$ & .098       &.165 & .197 &.166      & .193  &    .078   & .182  &  .139 & .170  & .109  \\
\hline
$f(L[0])$  & .260       &.430 & .437 & .407     & .362    &  .133   & .466  & .260 & .403 & .270 \\
\hline
$f(L)$  & .261 & .454      & .479 &  .441    & .456   &  .160    & .582 &  .272    & .409 & .278 \\
\hline
\end{tabular}
\vfill
\normalsize
\end{figure*}

We perform experiments on imagery from 3 different datasets. We refine the Stanford Background Dataset \cite{GouFulKol09} labeling to include a \texttt{vehicle} class with instance labeling. We also perform experiments on PASCAL VOC 2012 (Fig.~\ref{fig:voc2012} and Tables~\ref{tab:sum_iu_sbd} and \ref{tab:voc2012}). This dataset possesses relatively few images containing adjacent and/or overlapping instances of the same class. Therefore, we created a subset of the LM+Sun dataset \cite{tighe2010superparsing} of images containing at least 2 adjacent cars, consisting of 1,042 images.

\subsection{Features}\label{sec:features}
As discussed in \ref{sec:list_learning}, the features $\Phi(c,L)$ should encode the \emph{quality} of a chunk $c$ (e.g. ``Does the chunk look like a vehicle?'') and \emph{similarity} with the currently predicted list $L$ (e.g. ``Is this chunk similar to previously predicted chunks'' ). 
One of the quality features is built upon the superpixel-wise
multi-class label distribution from \cite{munoz-eccv-10}, where we
compute label distribution for each chunk via aggregating histograms
of its constituent superpixels. The other quality features are shape
features including central moments of the chunk, area, and scale
relative to the image. The similarity features we use primarily encode
spatial information between predictions. We use a candidate's
$\frac{I}{U}$ with previous predictions, the spatial histogram used in
\cite{desai2011discriminative} and the size of the current list. Chunks with high similarity
with previously predicted chunks in the list are less favored.

The features $\theta(s,c')$ for the grower encode information about the quality of proposed chunk $c' = c \cup \{s\}$ by growing $c$ with superpixel $s$. The grower uses the same quality features that characterize $c'$ used by the list predictor, as well as several of the class-agnostic features described in \cite{uijlings2013selective}, specifically color similarity (color histogram intersection), which encourages regions to be built from similarly colored regions, and region fill, which encourages growing compact chunks. See \cite{uijlings2013selective} for further details. As each superpixel is iteratively added to the chunk, similarity to the growing chunk for remaining candidate superpixels is recomputed. 

\begin{figure*}
\centering
\begin{minipage}[t]{\textwidth}
         \captionof{table}{List prediction and baseline performance on VOC2012 Person validation data and an 80/20 split of SBD Vehicles. Our list prediction outperforms all baselines in both experiments. In SBD Vehicles, the most competitive is the Scene Parsing intersected with DPM Bounding boxes. In VOC2012 Person, scene parsing was lower quality, and resulted in the DPM Boxes outperforming other baselines.} \label{tab:sum_iu_sbd}
         \begin{tabular}{|l||c|c|c|c|c||c|c|c|}
\cline{2-9}
\multicolumn{1}{c||}{} & \multicolumn{5}{c||}{SBD Vehicle} & \multicolumn{3}{c|}{VOC2012 Person} \\
\cline{2-9}
\multicolumn{1}{c||}{} & $f_{\text{L[0]}}$ & $f_{\text{L[0:1]}}$ & $f_{\text{L[0:2]}}$ & $f_{\text{L[0:3]}}$ & $f_{\text{L[0:4]}}$ &$f_{\text{L[0]}}$ & $f_{\text{L[0:1]}}$ & $f_{\text{L[0:2]}}$ \\
\hline
$\overline{R(c^*)}$ (mean optimal chunks given segmentation) & 0.82 & 1.43 & 1.87 & 2.19 & 2.44 &  0.83 & 1.17 & 1.33 \\
\hline
$\overline{R(c_\mathcal{G}^*)}$ (mean best grown chunks) & 0.69 & 1.14 & 1.45 & 1.66 & 1.81 &  0.52 & 0.71 & 0.79 \\
\hline
\multicolumn{9}{c}{}\\[-10pt]
\hline
List Prediction with Grown and Baseline Chunks & \textbf{0.58} & \textbf{0.89} & \textbf{1.08} & \textbf{1.18} & \textbf{1.25}& \textbf{0.38} & \textbf{0.50} & \textbf{0.53} \\
\hline
List Prediction with Selective Search & - & - & - & - & - &  0.27 & 0.36 & 0.41 \\
\hline
Scene Parsing $\cap$ DPM Baseline & 0.56 & 0.79 & 0.91 & 1.02 & 1.07  & 0.16 & 0.19 & 0.21 \\
\hline
Connected Components Baseline & 0.37 & 0.53 & 0.60 & 0.65 & 0.66 & 0.19 & 0.24 & 0.27 \\
\hline
DPM Baseline & 0.28 & 0.39 & 0.43 & 0.45 & 0.47 &  0.29 & 0.38 & 0.41 \\
\hline
\end{tabular}
\end{minipage}
\end{figure*}

\newlength{\widthlen}
\setlength{\widthlen}{.10\textwidth}
\begin{figure*}[!ht]
\centering
\begin{subfigure}[t]{\widthlen}
\includegraphics[width = \widthlen]{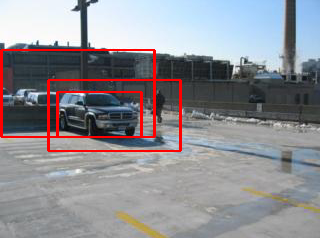}\caption*{DPM}
\end{subfigure}
\begin{subfigure}[t]{\widthlen}
\includegraphics[width = \widthlen]{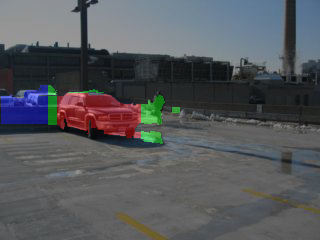}\caption*{SP $\cap$ DPM}
\end{subfigure}
\begin{subfigure}[t]{\widthlen}
\includegraphics[width = \widthlen]{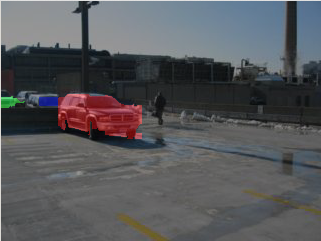}\caption*{Chunking}
\end{subfigure}
\hfill
\begin{subfigure}[t]{\widthlen}
\includegraphics[width = \widthlen]{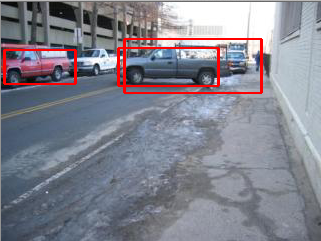} \caption*{DPM}
\end{subfigure}
\begin{subfigure}[t]{\widthlen}
\includegraphics[width = \widthlen]{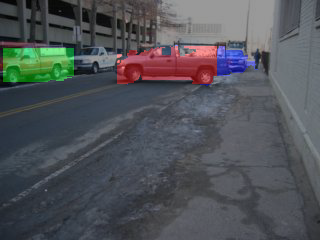} \caption*{SP $\cap$ DPM}
\end{subfigure}
\begin{subfigure}[t]{\widthlen}
\includegraphics[width = \widthlen]{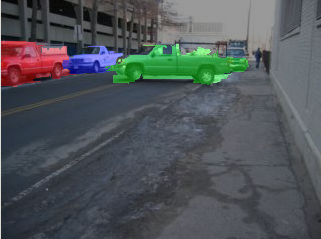}\caption*{Chunking}
\end{subfigure}
\hfill
\begin{subfigure}[t]{\widthlen}
\includegraphics[width= \widthlen]{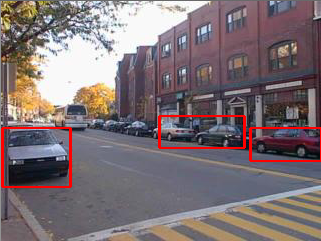} \caption*{DPM}
\end{subfigure}
\begin{subfigure}[t]{\widthlen}
\includegraphics[width= \widthlen]{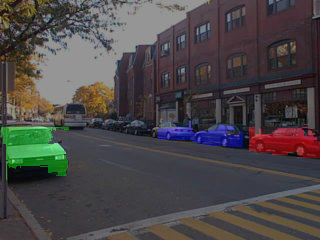}\caption*{SP $\cap$ DPM}
\end{subfigure}
\begin{subfigure}[t]{\widthlen}
\includegraphics[width= \widthlen]{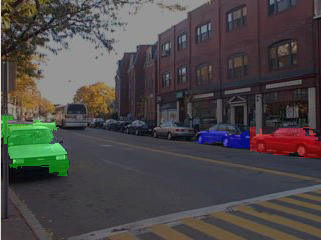}\caption*{Chunking}
\end{subfigure}
\caption{Comparison of list prediction versus other baselines. Each group of images contains, from left to right, the results of DPM, DPM intersected with Scene Parsing, and Visual Chunking. Note that while the Scene Parsing intersected with a bounding box detector can perform well, it fails in the case of poor NMS performance (group 1), and requires highly accurate Scene Parsing. Visual Chunking outperforms this baseline by instead building a list of detections.}
\label{fig:us_vs_them_all}
\end{figure*}

We evaluate three methods\footnote{
We use the semantic labeling algorithm
of \cite{munoz-eccv-10} and the DPM detection method of
\cite{felzenszwalb2010object} for bounding box output, with the
default SVM threshold, and NMS threshold of $0.5$. To generate the superpixels, we use the segmentation algorithm of \cite{felzenszwalb2004efficient}. For each experiment, separate
semantic labeling systems and chunk growers were trained. 
} leveraging existing bounding box detections and superpixel-wise semantic labeling algorithm, all of which serve as our baseline systems for building lists of predictions: \textbf{1)} Bounding box detector output after NMS filtering \textbf{2)} Connected components of scene parsing (``SP'') / semantic labeling \textbf{3)} A combination of \textbf{1)} and \textbf{2)}: intersection of connected components with bounding boxes, which creates chunks for every bounding box by extracting the labeled region inside (``SP $\cap$ DPM''). The third baseline is intended to capitalize on desirable properties of each component while avoiding their less desirable properties: boxes usually violate the object boundaries, and semantic labeling does not separate adjacent instances. The downside to this baseline is that it can suffer from
compounding both detector and scene parsing  errors. See Fig.~\ref{fig:us_vs_them_all} for a visual comparison.
We investigate the region generating methods of SCALPEL \cite{weiss2013scalpel} and Selective Search \cite{uijlings2013selective}, and in Fig.~\ref{fig:lmsun_abo_and_lp} compare our chunk generating method against them on our LM+Sun Adjacent Cars dataset with the Average Best Object method suggested by \cite{weiss2013scalpel}, and additionally train our system by using these methods to fill the candidate pool. In Table~\ref{tab:sum_iu_sbd}, we compare different list predictions methods on \texttt{vehicle} and \texttt{person} data, respectively.


\section{Conclusion}
We provide a novel method for producing region-based object detections
in images, treating the problem as a list prediction from a set of candidate
region proposals. We formulate a scoring criterion for multiple
object instances and multiple predictions. We develop a list
prediction algorithm that directly optimized the criterion. Our
approach is agnostic to proposal generation method and provides a recursive solution for all list lengths, enabling it to easily produce
any $k$ best guesses for objects. We provide a method for \textit{class-specific} candidate generation algorithm, yielding good coverage of objects. We demonstrate that our list prediction is a useful method for improving arbitrary candidate pools.


\bibliographystyle{IEEEtran}
\bibliography{root_flat}

\cleardoublepage
\section{Appendix}
This appendix contains proofs of theoretical results and additional pseudo-code descriptions presented in the paper.

\subsection{Proof for Theorem 1}
Given an image with ground truth entities $G = \{g_1, g_2, \dots, g_m\}$, candidate chunks set $C = \{c_1, c_2, \dots, c_n\}$ and list size budget $k$, our goal is to select the optimal $k$ chunks out of $C$ and associate each with the ground truth entities so as to maximize the sum of intersection over union scores under the association. Such problem can be cast as maximum weighted bi-partite graph matching, a classic assignment problem in combinatorial optimization. The edge set $E$ of the bi-partite graph is the Cartesian product of $G$ and $C$, i.e., $E = C \times G$. The weight $w_{ij}$ for each edge $e_{ij}$ is the I/U score between chunk $c_i$ and ground truth $g_j$. The defined quality metric $f(L,G)$ in Section 3.1 is equal to the optimal assignment score for subgraph $L \times G$.

 Let $V^* \subseteq C$ be the optimal size-$k$ subset of chunks, which can be computed in cubic time by Hungarian Algorithm~\cite{munkres1957algorithms}. Algorithm 1 can be seen as a greedy approach for maximum bi-partite graph matching with 1/2 approximation guarantee~\cite{preis1999linear}. Furthur, let $L_{gr}(V)$ be the greedy match on graph $V \times G$, we can show that for any augmented graph $V' \times G$ where $V \subseteq V'$, $L_{gr}(V')$ obtained from running Algorithm 1 with $k$ iterations is no worse than $L_{gr}(V)$. Hence, we can conclude that running Algorithm 1 on $C \times G$ has 1/2 approximation gurantee with respect to the optimal size-$k$ subset of $C$. Together with the fact that greedy solution has recursive structure, i.e., shorter greedy list is the prefix list for longer greedy list under larger budget, we can prove Theorem 1.     

\subsection{Proof of Theorem 2}
Consider superpixel $s_i$ and ground truth $g$ , let $\Delta_{x_i} = \|s_i \cap g\|$, $\Delta_{y_i} =\|s_i \cup g \| - \| g\|$ and $r_i = \frac{\Delta_{x_i}}{\Delta_{y_i}}$, we have that $\alpha_i = \frac{\Delta{x_i}}{s_i}$ is a montonic transformation of $r_i$, i.e., $r_i \geq r_j$ if and only if $\alpha_i \geq \alpha_j$. Therefore the rankings based on $\alpha_i$ or $r_i$ are the same. This follows from the fact that $\frac{1}{\alpha_i} - 1 = \frac{1}{r_i}$. 

Using the fact that superpixels are \textbf{non-overlapping}, given
any superpixel $s_i$ and a set of superpixel $c$, we have $R(c
\cup\{s_i\}) = \frac{\cap(c, g) + \Delta_{x_i}}{\cup(c, g) +
  \Delta_{y_i}}$. Further, if $r_i \geq R(c)$, adding $s_i$ to $c$
would increase $R(c)$ and vice versa, since $r_i =
\frac{\Delta_{x_i}}{\Delta_{y_i}} > \frac{\cap(c, g)}{\cup(c, g)} $
implies $ r_i > R(c\cup\{s_i\}) > R(c) =  \frac{\cap(c, g)}{\cup(c,
  g)} $. Therefore, suppose the optimal solution be $c^*$, then if
$r_i > R(c^*)$, it must be true that $s_i \in c^*$, and otherwise $s_i
\notin c^*$. 
This also implies the optimal set of superpixels is the first $k$ elements based on a sorting of superpixels by $r_i$, where $k$ is the smallest integer such that $r_{k+1} \leq R(c^*)$. 



\subsection{Proof of Theorem 3 and Corollary}
Let $\alpha_1 \geq \alpha_2 \geq \cdots \geq \alpha_N $ and let the
optimal set of superpixels to maximize I/U with ground truth $g$ be the first $k$ superpixels, i.e., $c^* = \{
s_1, s_2, \dots, s_k \}$; suppose the regressor makes bounded uniform
error $\epsilon$, i.e., $|\hat{\alpha}_i - \alpha_i| < \epsilon$, and let $M$ be the largest number such that: $\alpha_j \geq \alpha_k - 2\epsilon$ for $j = 1 ,..., M$. 
If the regressor makes bounded uniform error $\epsilon$, then the
worst case would be: it underestimates $\alpha_1 ,..., \alpha_k$ by
$\epsilon$ and overestimates $\alpha_{k+1} ,..., \alpha_M $ by
$\epsilon$. Therefore, some of the elements in $\alpha_{k+1} ,...,
\alpha_M $ would rank higher than elements in $\alpha_1 ,..., \alpha_k$.  
Denote $c_{\mathcal{G}}^*$ as the best solution among the chain of sets induced by the ranking output of the regressor $\mathcal{G}$.  
\begin{align}
  R(c_{\mathcal{G}^*}) & \geq R(\{s_1, ..., s_M \}) = \frac{\sum_{i=1}^M{s_i \alpha_i}}{g + \sum_{i=1}^M{s_i (1-\alpha_i)}} \\
  & = \frac{\sum_{i=1}^k{s_i \alpha_i} + \sum_{j=k+1}^M{s_j \alpha_j} }{g + \sum_{i=1}^k{s_i (1-\alpha_i)} + \sum_{j=k+1}^M{s_j (1-\alpha_j)}} \\
  & \geq \min\{ \frac{\sum_{i=1}^k{s_i \alpha_i}}{g+ \sum_{i=1}^k{s_i (1-\alpha_i)}},  \frac{\sum_{i=k+1}^M{s_i \alpha_i}}{\sum_{i=k+1}^M{s_i (1-\alpha_i)}} \} \\
  & = \min\{ R(c^*), \frac{\sum_{i=k+1}^M{s_i \alpha_i}}{\sum_{i=k+1}^M{s_i (1-\alpha_i)}}  \} \\
  & = \frac{\sum_{i=k+1}^M{s_i \alpha_i}}{\sum_{i=k+1}^M{s_i (1-\alpha_i)}}   \geq \frac{\sum_{i=k+1}^M{s_i (\alpha_k - 2\epsilon) }}{\sum_{i=k+1}^M{s_i (1-(\alpha_k -2\epsilon) )}} \\
 & = \frac{ (\alpha_k - 2\epsilon)} {(1-\alpha_k + 2\epsilon )  } \geq \frac{R(c^*)(\alpha_k - 2\epsilon)}{\alpha_k + 2\epsilon R(c^*)}  \label{moveOptin}
\end{align}
In \ref{moveOptin}, we are using the fact that  $r_k = \frac{\alpha_k}{1 - \alpha_k} \geq R(c^*)$ implies $ 1 - \alpha_k \leq \frac{\alpha_k}{R(c^*)}$. Rearrange the terms, we get:
\begin{align}
\frac{R(c)} {R(c^*)} & \geq \frac{\alpha_k - 2\epsilon}{\alpha_k + 2\epsilon R(c^*)} 
 \geq 1 -  \frac{ 2(1 + R(c^*))}{\alpha_k} \epsilon \label {beforerelaxfracalpha}\\ 
& \geq 1 - \frac{4}{(1/R(c^*) + 1)} \epsilon \label{afterrelaxfracalpha}
\end{align}

From \eqref{beforerelaxfracalpha} to \eqref{afterrelaxfracalpha}, we are using the fact that $\frac{1}{r_k} = \frac {1}{\alpha_k} - 1 \leq \frac{1}{R(c^*)}$ and $R(c^*) \leq 1$.  
We can proceed to have an additive bound:
\begin{align}
R(c^*) - R(c) &\leq \frac{4R(c^*)} {1/R(c) + 1} \epsilon \label{beforerelaxf(OPT)}  \\ 
                  & \leq 2\epsilon \label{afterrelaxf(OPT)}
\end{align}
A more natural assumption is to assume an expected square error $\epsilon$ over the distribution $P$ of all superpixels. Denote $\delta = \E_{i \sim P} (\hat{r_i} - r_i)^2$, the expected uniform error bound $\E[\epsilon]$ satisfies:
\begin{align}
 \E[\epsilon] & = \E [ \max_i | \epsilon_i | ] = \E [ ( \max_i \epsilon_i^2 )^{\frac{1}{2}} ] \\
 & \leq \E [ ( \sum_i \epsilon_i^2 )^{\frac{1}{2}} ]  
 \leq \E[ n \epsilon^2 ]^{1/2}  \label{afterJensen} \\
&\leq \sqrt{n \delta}
\end{align}
In \eqref{afterJensen}, we are applying Jensen's Inequality along with the fact that $\sqrt{x}$ is concave.
Using Markov Inequality, for any $\eta \in (0,1)$, with probability $1 - \eta$, we have that: 
\begin{align}
\epsilon \leq \frac{\sqrt{n\delta}}{\eta}
\end{align}
Together with \ref{afterrelaxf(OPT)}, we have that: for any $\eta \in (0,1)$, with probability $1 - \eta$
\begin{align}
R(c_{\mathcal{G}}^*) > R(c^*) - 2\frac{\sqrt{n\delta}}{\eta}
\end{align}

\subsection{Pseudocode for Multiple Grower Algorithm}
\begin{algorithm}[hb!] 
\caption{Multiple Instance Chunk Growing Algorithm\label{Grower_Multi}}
\begin{algorithmic}
\State \textbf{Input:} Set of superpixels $S$, grower predictor $\mathcal{G}$, seeding superpixel $s'$ 
\State \textbf{Output:} A set of chunks $C_{\mathcal{G}}$.
\State $c = \{s'\}$, $C_{\mathcal{G}} = \emptyset$
\For{$i=1$ \textbf{to} $|S|$}
\State $s_i = \argmax_{s\in S}\mathcal{G}(s, c) $
\State $c = c \cup\{s_i\}$, $C_{\mathcal{G}} = C_{\mathcal{G}} \cup \{c\}$
\EndFor
\State Return $C_{\mathcal{G}}$
\end{algorithmic}
\end{algorithm}

The pseudocode in Algorithm~\ref{Grower_Multi} describes the growing algorithm
with specified seeding superpixel $s'$. We run this algorithm for each
$s'\in L$ in order to increase diversity, where $L$ is the set of seeding superpixels.
Two major differences from single instance chunk growing algorithm in Section 3.2 are addressed below:
\begin{enumerate}
\item A seeding superpixel $s'$ needs to be given as input to initialize the chain of growth, i.e., $c = \{ s' \}$.
\item Instead of just using features only based on the superpixel $s$ itself, we also consider features including both the superpixel and the currently growing chunk $c$. We replace $\hat{\alpha_i} = \mathcal{G}(s_i)$ with $\hat{\alpha_i} = \mathcal{G}(s_i, c)$. These feature not only encode information about the quality of a superpixel but also encourage the grower to grow spatially compact chunks.   
\end{enumerate}

\end{document}